\documentclass[sigconf]{acmart}
\AtBeginDocument{%
  }

\setcopyright{acmlicensed}
\copyrightyear{2025}
\acmYear{2025}
\acmDOI{XXXXXXX.XXXXXXX}
\acmConference[MM'25]{Make sure to enter the correct
  conference title from your rights confirmation email}{October 27–31, 2025}{Dublin, Ireland}
\acmBooktitle{In Proceedings of the 33rd ACM International Conference on Multimedia (MM’25), October 27-31, 2025, Dublin, Ireland}
\acmISBN{978-1-4503-XXXX-X/2018/06}

\usepackage{amsmath}
\usepackage{subcaption}
\usepackage{todonotes}
\usepackage{multirow}
\usepackage{colortbl}
\usepackage{array}
\definecolor{colorTrd}{rgb}{0.65,0.95,0.95}
\definecolor{colorSnd}{rgb}{0.7, 0.85, 1}
\definecolor{colorFst}{rgb}{0.7, 0.7, 1}
\usepackage{colortbl}

\usepackage{amsfonts}
\usepackage{float}
\usepackage{multicol}
\usepackage{booktabs}

\usepackage{pifont}

\usepackage[linesnumbered,ruled,vlined]{algorithm2e}

\begin{document}

\title{DenseSR: Image Shadow Removal as Dense Prediction}

\author{Yu-Fan Lin*}
\affiliation{%
  \institution{National Cheng Kung University}
  \city{Tainan}
  \country{Taiwan}
}
\email{aas12as12as12tw@gmail.com}

\author{Chia-Ming Lee}
\authornote{Equal contribution.}
\affiliation{%
  \institution{National Cheng Kung University}
  \city{Tainan}
  \country{Taiwan}
}
\email{zuw408421476@gmail.com}

\author{Chih-Chung Hsu}
\authornote{Corresponding author.}
\affiliation{%
  \institution{National Yang Ming Chiao Tung University}
  \city{Tainan}
  \country{Taiwan}
}
\email{chihchung@nycu.edu.tw}

\renewcommand{\shortauthors}{Lin et al.}



\begin{abstract}
Shadows are a common factor degrading image quality. Single-image shadow removal (SR), particularly under challenging indirect illumination, is hampered by non-uniform content degradation and inherent ambiguity. Consequently, traditional methods often fail to simultaneously recover intra-shadow details and maintain sharp boundaries, resulting in inconsistent restoration and blurring that negatively affect both downstream applications and the overall viewing experience. To overcome these limitations, we propose the \textbf{DenseSR}, approaching the problem from a dense prediction perspective to emphasize restoration quality. This framework uniquely synergizes two key strategies: (1) deep scene understanding guided by geometric-semantic priors to resolve ambiguity and implicitly localize shadows, and (2) high-fidelity restoration via a novel Dense Fusion Block (DFB) in the decoder. The DFB employs adaptive component processing—using an Adaptive Content Smoothing Module (ACSM) for consistent appearance and a Texture-Boundary Recuperation Module (TBRM) for fine textures and sharp boundaries—thereby directly tackling the inconsistent restoration and blurring issues. These purposefully processed components are effectively fused, yielding an optimized feature representation preserving both consistency and fidelity. Extensive experimental results demonstrate the merits of our approach over existing methods. Our code can be available on \href{https://github.com/VanLinLin/DenseSR}{\textcolor[rgb]{0.90, 0.19, 0.49}{https://github.com/VanLinLin/DenseSR}}.
\end{abstract}
\maketitle

\section{Introduction}
\label{sec:intro}

Shadows, as natural consequences of light-object interactions, are ubiquitous optical phenomena in the visual world. The presence of shadows profoundly impacts multimedia content analysis, degrading performance in tasks ranging from remote sensing \cite{rtcs}, segmentation \cite{10678598}, tracking \cite{5597618} and 3D reconstruction \cite{Weder_2023_CVPR,bolanos2024gsc} to multimedia applications \cite{9918057}. 
Removing shadows from images to restore the authentic appearance of occluded regions is not only a fundamental computer vision task but also a critical step for enhancing downstream application performance \cite{8378149,murali2016survey,tiwari2016survey,MM}. The core challenge of this task lies in accurately understanding the local illumination attenuation patterns (distinguishing shadows from intrinsic object darkness) and leveraging contextual information to perform physically plausible and visually natural content filling and color correction within shadowed areas for restoration \cite{le2019shadow,yang2012shadow}.

\begin{figure}[t!]
\centering
\includegraphics[width=1.0\linewidth]{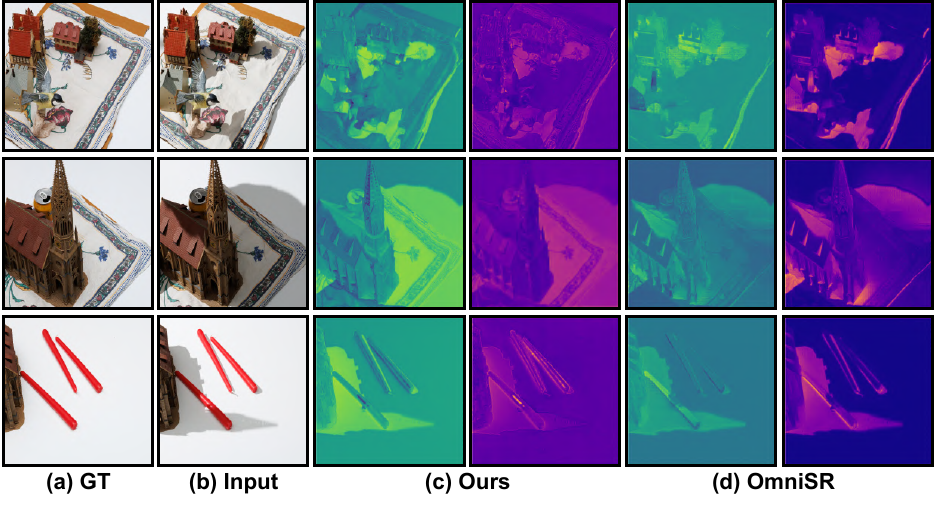}
\vspace{-0.7cm}
\caption{To tackle inconsistent restoration and boundary blurring in shadow removal, we employs an adaptive strategy. As illustrated, it distinctly processes: (Left) smoothed base features ensuring content consistency (akin to mean, processed via ACSM smoothing); and (Right) high-frequency features for detail recovery and boundary sharpening (akin to variance, refined via TBRM). Fusing these purposefully processed features enables high-quality shadow removal that balances both content consistency and boundary clarity.}
\label{fig:first} 
\vspace{-0.2cm}
\end{figure}
Despite significant advances driven by deep learning in single-image shadow removal, several deep-seated bottlenecks remain. \textbf{First}, the ambiguity between shadows and intrinsic object properties remains challenging to resolve solely based on RGB information. \textbf{Second}, the complexity of real-world illumination, particularly the prevalence of indirect lighting and the resulting soft shadows in indoor scenes, is often inadequately addressed, limiting model performance in such scenarios, partly due to insufficient modeling of physical light transport like scattering and diffusion. \textbf{Third}, standard feature fusion strategies employed in hierarchical networks exhibit inherent flaws: they typically assume features accurately represent scene content at their respective scales. However, shadows non-uniformly degrade this representation, causing simple fusion methods to fail in handling this spatially varying signal degradation, resulting in inconsistent intra-shadow restoration and significant loss of boundary details.

To overcome these bottlenecks, our approach returns to the physical essence of shadow formation and fundamental principles of information processing. As revealed by fundamental shading models, object appearance results from a complex interplay of illumination, geometry (surface orientation), and material (reflectance properties). Shadows fundamentally alter the illumination component. Accurately inverting this effect to obtain the shadow-free image necessitates effectively disentangling illumination effects from intrinsic properties, strongly motivating the incorporation of external prior knowledge capturing geometry and material/semantic characteristics. Concurrently, recognizing the failure of standard fusion strategies when dealing with shadow-degraded features, we identified the need for a more sophisticated and adaptive fusion mechanism. Such a mechanism must be capable of distinguishing and processing different information components affected by shadows—for instance, the relatively stable low-frequency base appearance versus the heavily distorted or obscured high-frequency texture details.

Based on these motivations, we propose the DenseSR framework, approaching shadow removal from a dense pixel-wise prediction perspective. The core of DenseSR lies in a two-parts: first, it achieves deep scene understanding and implicit shadow localization/disambiguation by integrating powerful geometric (depth, normal) and semantic (DINO) priors guided through attention mechanisms in Scene-Integrated Modules (SIM); building upon this understanding, we introduce the innovative Dense Fusion Block (DFB) within the decoder, specifically responsible for high-fidelity content restoration. DFB employs an adaptive component processing approach: the Adaptive Content Smoothing Module (ACSM) focuses on restoring a consistent base appearance within the shadow region from coarser-scale features, suppressing noise and artifacts; meanwhile, the Texture-Boundary Recuperation Module (TBRM) concentrates on recuperating obscured fine textures and sharpening boundaries using finer-scale features, as shown in Figures \ref{fig:first} and \ref{fig:evolution}. These complementary modules yield effectively combined outputs, generating an optimized feature representation that preserves both internal consistency and boundary details, ultimately enabling high-quality shadow removal. The main contributions of this study can be summarized into three points:

\begin{itemize}
    \item A novel shadow removal framework (DenseSR) integrating prior knowledge: Approaching the task from a dense prediction perspective, this framework utilizes attention mechanisms to effectively guide geometric and semantic priors, addressing the core shadow ambiguity issue. 
    \item The design of a Dense Fusion Block (DFB) tailored for shadow degradation: Featuring complementary ACSM and TBRM modules, its adaptive component processing strategy specifically targets the intra-shadow inconsistency and boundary/detail loss issues characteristic of standard fusion methods in shadow removal.
    \item Demonstration of state-of-the-art performance under complex illumination: Extensive experiments validate DenseSR's robustness and effectiveness, particularly in handling challenging direct and indirect illumination scenarios. 
\end{itemize}
The following sections will detail related work, motivation, network architecture, experimental setup, and results analysis.

\section{Related Work}
\label{sec:related}

\subsection{Single Image Shadow Removal}

Single-image shadow removal aims to restore the authentic appearance beneath shadows, a fundamental computer vision task \cite{8378149,murali2016survey,tiwari2016survey,guo2024single}. Early traditional single-image shadow removal methods typically operated in two stages: first detecting shadow regions, then performing removal. These techniques relied heavily on handcrafted features, physical or statistical models, and strong assumptions about illumination and surfaces \cite{shor2008shadow,zhang2015shadow,cucchiara2003detecting,salamati2011removing}. The removal stage employed physics-inspired strategies, such as image decomposition into illumination and reflectance components \cite{Le_2019_ICCV}. However, their reliance on specific priors and heuristic models made them struggle with complex scenes, soft shadows, and varying conditions, thus limiting their generalization capability.

Fortunately, deep learning significantly advanced the field: CNNs \cite{ronneberger2015u, qu2017deshadownet} captured multi-scale features but faced locality limits; Transformers \cite{shadowformer, Xiao_2024_CVPR, RASM} offered better global context, yet ambiguity persisted without priors \cite{omnisr}, and some relied on masks \cite{shadowformer}; Diffusion models \cite{guo2023shadowdiffusion, mei2024latent} achieve high quality at significant computational cost. Early reliance on masks \cite{Mask_ShadowNet} simplified learning but proved impractical, motivating mask-free approaches that must jointly locate and restore shadows \cite{shadowrefiner}. While incorporating priors aids ambiguity resolution, existing methods often neglect complex light physics (e.g., scattering, diffusion shading), hindering adaptive restoration for diverse shadow types and origins (esp. indirect light) and causing boundary smoothing or internal artifacts. Furthermore, standard feature fusion in hierarchical networks degrades restoration quality, losing boundary details and improperly mixing intra-object features, leading to inconsistency within recovered shadows.

\subsection{Dense Prediction}
Dense prediction tasks form a core category of problems in computer vision, aiming to predict a corresponding value for every pixel in an input image. This encompasses a wide range of applications such as semantic segmentation \cite{long2015fully,ronneberger2015u}, instance segmentation \cite{8237584}, object detection \cite{carion2020DETR}, and image restoration/translation tasks like shadow removal.
Foundational architectures FCN \cite{long2015fully} and U-Net \cite{ronneberger2015u} established hierarchical designs to capture multi-scale information. Subsequently, architectures incorporating FPN \cite{8099589} became widely adopted for many dense prediction tasks, explicitly providing features at multiple resolutions.

A critical component in these hierarchical and FPN-like structures \cite{denseprediction,multipathdense,wang2021pyramid} is effective \textbf{feature fusion}. Since deep layers in these networks capture coarse, high-level semantic information while shallow layers retain fine-grained, high-resolution spatial details, fusion is essential to combine these complementary representations for generating high-quality, high-resolution predictions \cite{Hsu_2024_CVPR,goshtasby2005fusion}. However, effectively fusing features across significant gaps in resolution and semantic levels remains a key challenge. Simple fusion strategies, such as upsampling (e.g., via bilinear interpolation) followed by element-wise addition or concatenation, often struggle to adequately integrate information from different scales. This frequently leads to the loss of crucial high-frequency details, potentially resulting in inconsistency within predicted regions (intra-category inconsistency), blurred object boundaries, or other artifacts. Consequently, advanced fusion techniques have been developed, employing strategies like adaptive kernels \cite{10377871,wang2019carafe}, enhanced interactions \cite{lu2022sapa}, or feature alignment \cite{huang2021alignseg} to improve detail preservation and adaptability. 

While shadow removal is a dense prediction task plagued by issues like internal inconsistency and boundary blurring often exacerbated by simplistic feature fusion \cite{omnisr, shadowformer, shadowrefiner}, many current approaches have not fully adopted or specifically adapted the more sophisticated fusion techniques required to effectively address the non-uniform feature degradation unique to shadows.
\section{Preliminary and Motivation}
\label{sec:prelim_motivation}

\begin{figure}[t!]
\centering
\includegraphics[width=1.0\linewidth]{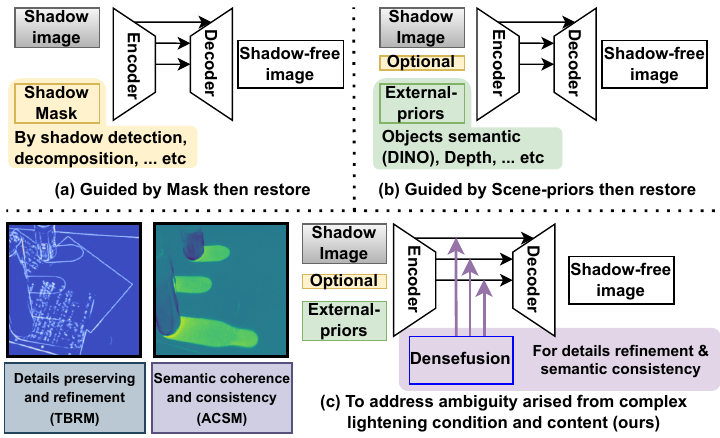}
\vspace{-0.4cm}
\caption{Evolution of shadow removal approaches. (a) Mask-guided. (b) Prior-guided. (c) Our method (DenseSR) combines scene-priors guidance with the proposed Dense Fusion strategy. The Dense Fusion tackles the non-uniform feature degradation caused by shadows, aiming to better preserve details and ensure semantic coherence during restoration.}
\vspace{-0.2cm}
\label{fig:evolution} 
\end{figure}
\begin{figure*}[t!]
\centering
\includegraphics[width=1.0\linewidth]{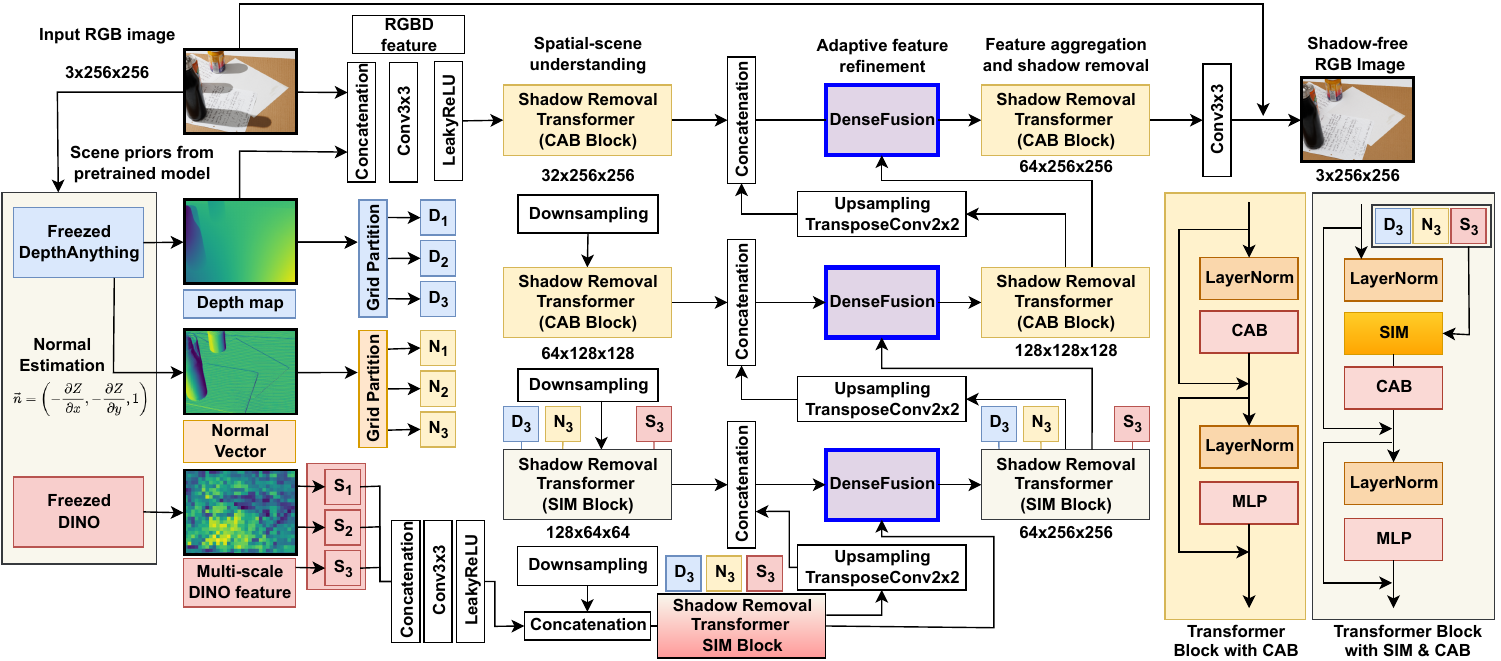}
\vspace{-0.7cm}
\caption{Overall architecture of the proposed DenseSR framework, embodying a layered 'understand then restore' strategy. The encoder leverages multi-modal scene priors (Depth, Normal, DINO features extracted via pre-trained models), integrated through standard Swin-Transformer with SIM blocks for ambiguity resolution. The symmetric decoder path employs DFB after each upsampling and skip connection stage to perform adaptive, high-fidelity feature restoration before the final output projection reconstructs the shadow-free image. Note that {${D_{i}}$,${N_{i}}$,${S_{i}}$} denote the corresponding feature with different scale to match the feature map resolutions at different stages, larger $i$ values represent smaller sizes.}
\label{fig:flowchart} 
\vspace{-0.2cm}
\end{figure*}

\subsection{Shadow Physics, Image Model, and Challenges}
The formation of shadows originates from the fundamental physical principle of light propagation being occluded by 3D scene geometry. The final appearance of any point in an image is determined by the aggregation of all light rays arriving at that point, interacting with the surface material (described by the BRDF), and scattering towards the viewing direction. Incident illumination can be conceptually divided into direct illumination from primary sources and indirect illumination resulting from scene reflections/scattering. Occlusion of direct light creates well-defined shadows, whereas occlusion of indirect light—pervasive in indoor environments rich with complex light interactions like ambient light and interreflections—forms softer, graded shadows, the accurate modeling of which is crucial for realistic restoration.

At the image level, shadows manifest as local attenuation in brightness and potential color shifts compared to the shadow-free state. While often simplified using a multiplicative model $\mathbf{I}_s(x) \approx \mathbf{I}_f(x) \times \mathbf{A}(x)$ (where $\mathbf{A}(x)$ is a spatially varying illumination factor), the true impact is more complex, involving non-linear effects and spatial/directional variations in illumination. Consequently, recovering the shadow-free image $\mathbf{I}_f$ from a single shadowed observation $\mathbf{I}_s$ is a highly challenging ill-posed inverse problem. Core challenges include: 1) \textbf{Ambiguity}: The visual similarity between shadows and intrinsically dark surfaces. 2) \textbf{Complex Lighting Physics}: Difficulty in accurately modeling indirect illumination and the resulting soft, graded shadows. 3) \textbf{Non-uniform Feature Degradation}: Shadows impact image content non-uniformly across space, complicating subsequent processing.

\subsection{Evolution of Learning Strategies and Motivation for DenseSR}
With the development of learning strategies for shadow removal, methods have evolved to tackle these challenges. Early methods attempted to simplify the task using shadow masks $\mathbf{M}$ by shadow detection techniques and learning mapping functions $\mathcal{F}': (\mathbf{I}_s, \mathbf{M}) \mapsto \hat{\mathbf{I}}_f$, but the practical difficulty of obtaining masks limits applicability. Concretely, complex illumination and lighting conditions make shadow detection fail in various scenes. Thus, the mask-free setting $\mathcal{F}: \mathbf{I}_s \mapsto \hat{\mathbf{I}}_f$ became predominant, requiring the model to implicitly disentangle illumination attenuation from image content.

On the other hand, recent advancements in large pre-trained foundation models offer new avenues. The rich general visual knowledge (encompassing geometry, semantics, materials) learned by these models can be transferred. In shadow removal, researchers explore leveraging geometric and semantic priors derived from such models (e.g., DINO \cite{dinov2}, Depth Anything \cite{depthganythingv2}), inspired by attempts like \cite{omnisr, lee2025prompthsiuniversalhyperspectralimage}. These priors can provide crucial contextual cues to help mitigate the ambiguity between shadows and dark objects and implicitly infer regions likely requiring restoration, partially substituting the role of masks.

Therefore, our core motivation stems from the observation that while powerful priors aid scene understanding and localization, the primary bottleneck becomes high-fidelity content restoration. The high-level idea is briefly illustrated in Figures \ref{fig:first} and \ref{fig:evolution}. We thus regard the task as a dense prediction problem focused on restoration quality. However, because shadows cause non-uniform feature degradation, standard feature fusion mechanisms are inadequate during restoration, failing to simultaneously ensure content consistency and detail clarity. This necessitates advanced network modules capable of adapting to shadow-specific degradation characteristics (like our proposed DFB), specifically designed to address challenges within the restoration phase itself and generate high-quality pixel-level outputs. This forms the starting point for the DenseSR framework design.
\section{Proposed Method}

\begin{figure}[t]
\centering
\includegraphics[width=0.95\linewidth]{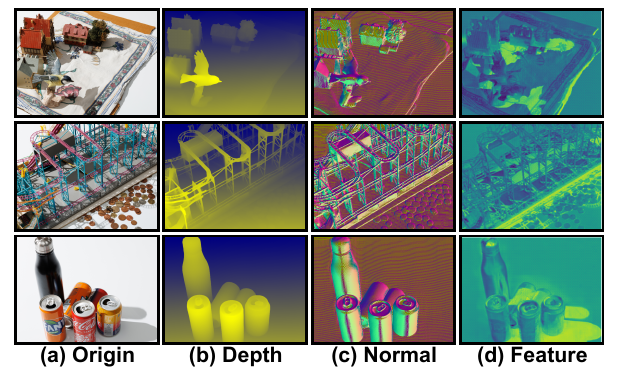}
\caption{Multi-modal scene priors and feature map within decoder, including the estimated depth map $\boldsymbol{D}$, capturing camera-to-point distances, and the derived surface normal map $\boldsymbol{N}$, indicating surface orientations. 
These geometric priors, along with semantic features, guide our network towards more physically plausible shadow removal.}
\vspace{-0.5cm}
\label{fig:modalities} 
\end{figure}

To address the inherent challenges of single-image shadow removal, particularly the demand for high-fidelity restoration within the dense prediction paradigm, we propose the DenseSR framework. Designed to learn a precise mapping $F: \mathbf{I}_s \mapsto \hat{\mathbf{I}}_f$ from a shadowed input $\mathbf{I}_s$ to its corresponding shadow-free output $\hat{\mathbf{I}}_f$, DenseSR follows a layered strategy: (1) leveraging multi-modal priors for deep scene understanding and context awareness in the encoder; and (2) subsequently executing precise, restoration-oriented feature fusion and enhancement via innovative Dense Fusion Blocks (DFBs) in the decoder. The overall network architecture is based on U-Net \cite{ronneberger2015u} and integrates Swin Transformer \cite{liu2021swin, wang2022uformer} attention mechanisms (see Figure \ref{fig:flowchart}).

\subsection{Network Architecture Overview}
\label{subsec:arch_overview}

DenseSR adopts a symmetric encoder-decoder design. The input first passes through a $3\times3$ Convolution layer with LeakyReLU mapping RGBD features to the initial embedding space. The encoder consists of three stages, each comprising consecutive Transformer Blocks and a $4\times4$ convolution with $2$ strides per-move for downsampling, progressively reducing spatial resolution while increasing channel dimensionality. A bottleneck layer, also using a Transformer Block, processes the deepest features. The decoder mirrors this structure with three corresponding stages, each including an $2\times2$ transposed convolution for upsampling, concatenation with features from the corresponding encoder level via skip connection, a core module DFBs, illustrated in Sec.~\ref{subsec:DFB}, and a Transformer Blocks. Finally, a $3\times3$ convolution layer maps the decoder output back to the RGB space, which is added to the original input $\mathbf{I}_s$ through a global residual connection to yield the final shadow-free image $\hat{\mathbf{I}}_f$.

\subsubsection{Spatial and Scene-prior Extraction and Preprocessing}
\label{subsec:prior_integration_encoding}

Accurate shadow removal hinges on the network's ability to understand the input scene $\mathbf{I}_s$ to distinguish illumination effects from intrinsic surface properties. We achieve this by incorporating powerful geometric and semantic priors. Utilizing pre-trained models Depth-Anything-V2 \cite{depthganythingv2} and DINO-V2 \cite{dinov2}, we extract a depth map $\mathbf{D}$ (from which the normal map $\mathbf{N}$ is derived by normal estimation \cite{bae2021eesnu}) and multi-scale DINO feature maps $\mathbf{F}_{multiscale}$, rich in material and high-level semantic information \cite{sharma2023materialistic}. The RGB image is concatenated with the depth map $\mathbf{D}$ to form the primary RGBD input $\mathbf{X}_{RGBD}$. The extracted priors (\{$\mathbf{D, N, F}\}_{multiscale}$) are preprocessed (e.g., via grid partition and sampling) to match the feature map resolutions at different stages of the encoder and bottleneck.

\subsubsection{Scene-prior-Modulated Attention Mechanism}
The deep fusion of prior information primarily occurs within the Transformer Blocks of the encoder and bottleneck, particularly in the deeper layers configured as Scene-Integrated Modules (SIM) . The core of these modules is a specially designed window attention. This attention mechanism goes beyond standard self-attention by introducing a prior-based modulation process:

\begin{itemize}
    \item Compute Similarity/Consistency Maps: Within each attention window, using the spatially corresponding prior information, it dynamically computes pairwise semantic similarity maps (based on DINO feature $\mathbf{F}$ dot-product correlations) and geometric consistency maps (based on planar distance calculations using depth $\mathbf{D}$ and normals $\mathbf{N}$).
    \item Modulate Attention Scores: The computed semantic similarity and geometric consistency maps are then used to element-wise modulate the standard $QK^T$ attention score map before the Softmax operation.
\end{itemize}

The significance of this mechanism lies in its ability to make the self-attention weights explicitly dependent on the geometric structure and semantic content of the local scene context. This allows the network to aggregate information more intelligently, for instance, by prioritizing interactions between pixels belonging to the same object surface or geometrically coherent regions. Consequently, it significantly mitigates the ambiguity between shadows and dark surfaces, a core challenge in shadow removal.
\begin{figure*}[t!]
\centering
\includegraphics[width=1.0\linewidth]{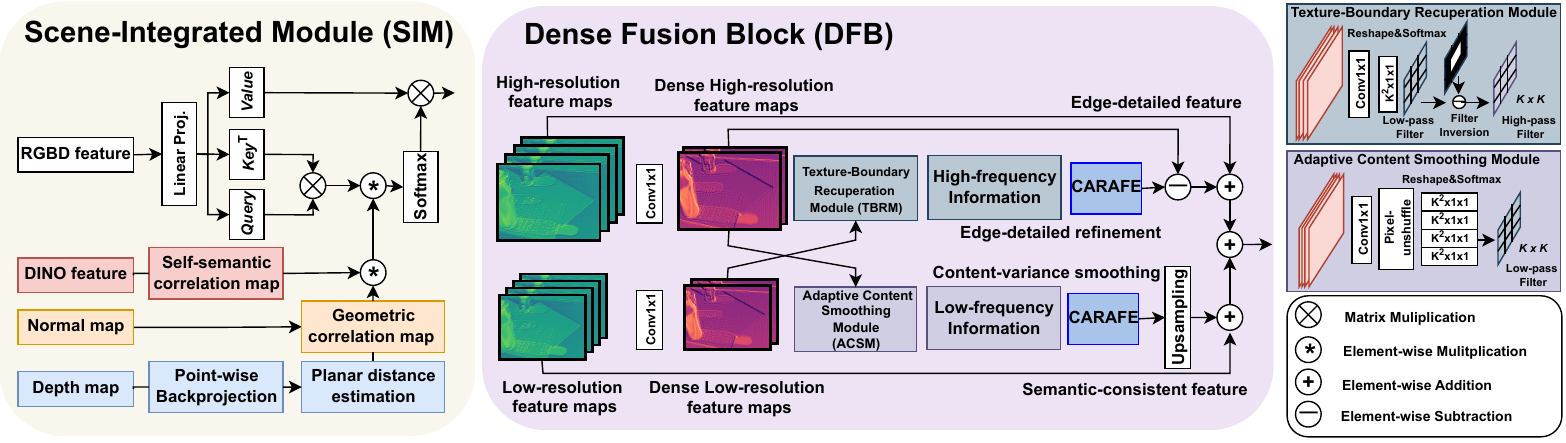}
\vspace{-0.7cm}
\caption{The SIM leverages geometric and semantic priors to make its self-attention context-aware, enhancing scene understanding and reducing shadow ambiguity. Subsequently, the DFB performs adaptive feature fusion with content-filtering manner to handle shadow degradation. Its ACSM sub-module ensures content consistency using adaptive smoothing filters, while its TBRM sub-module recovers fine textures and sharp boundaries using adaptive detail-enhancing filters. By fusing these complementary processed outputs, the DFB generates a high-quality feature representation that balances consistency and fidelity, outperforming previous methods and standard fusion for effective shadow removal.}
\label{fig:densefuse} 
\vspace{-0.2cm}
\end{figure*}

\subsubsection{Enhancement with Global Semantic Information}
At the network's deepest bottleneck layer, the use of global semantic information is further reinforced. Multi-scale DINO features are processed and integrated, concatenated with the deepest encoder output, and then jointly fed into the bottleneck's SIM block. This ensures that high-level semantic context fully informs the most abstract feature processing stage.

Through these prior integration and contextual encoding mechanisms, the DenseSR encoder generates multi-scale feature representations that are rich in context, exhibit reduced shadow ambiguity, and possess enhanced discriminability, providing a high-quality input foundation for the subsequent restoration task. By leveraging priors within the SIMs for scene understanding and ambiguity reduction, the encoder prepares context-aware features. This foundation enables the decoder's DFB to effectively concentrate on the demanding challenge of high-fidelity content restoration within the implicitly identified shadow regions. The design of the DFB will now be detailed.

\subsection{Dense Fusion for Shadow Removal}
\label{subsec:DFB}

Having understood the scene and partially resolved ambiguities using priors, the core restoration task is performed in the decoder. Because shadows induce complex and non-uniform degradation on image content, simple feature fusion struggles to meet the pixel-level accuracy demands of the dense prediction task. To this end, we introduce the DFB at each upsampling stage of the decoder, as shown in Figure \ref{fig:flowchart}, with its detailed structure shown in Figure \ref{fig:densefuse}. The DFB is designed specifically to counteract shadow-induced feature degradation and maximize content preservation and recovery during cross-scale information fusion. Concretely, DFB employs an adaptive component processing strategy, incorporating two functionally complementary core sub-modules:

\subsubsection{Adaptive Content Smoothing Module (ACSM)}

Hierarchical network architectures often suffer from intra-feature inconsistency when upsampling deep, low-resolution feature maps ($\mathbf{Y}^{l+1}$), particularly within shadow regions where illumination is uneven or original textures are obscured. To promote a consistent and uniform appearance restoration, the ACSM is employed. Its primary objective is to predict spatially-variant smoothing filters based on local contextual information ($\mathbf{Z}^l$), typically derived from an initial fusion of corresponding features. Specifically, the module first passes $\mathbf{Z}^l$ through a convolutional layer to estimate raw filter weights $\bar{\mathbf{V}}^l$ for each spatial location $(i,j)$. Subsequently, a channel-wise Softmax function is applied across the $\bar{K}^2$ dimension (where  
$\bar{K}$ is the kernel size) to yield normalized, adaptive smoothing filter kernels $\bar{\mathbf{W}}^l$:
\begin{equation}
\label{eq:acsm_softmax}
\bar{\mathbf{W}}_{i,j,k}^l = \text{Softmax}(\bar{\mathbf{V}}_{i,j,:}^l)_k = \frac{\exp(\bar{\mathbf{V}}_{i,j,k}^l)}{\sum_{k'=1}^{\bar{K}^2} \exp(\bar{\mathbf{V}}_{i,j,k'}^l)},
\end{equation}

where $k$ indexes the filter weights. These predicted adaptive filters  
$\bar{\mathbf{W}}^l$ are designed to gently blur high-frequency variations, thereby enhancing content consistency. Following the efficient implementation strategy inspired by \cite{10377871,wang2019carafe}, these filters can be applied concurrently with $2\times$ upsampling using mechanisms like Pixel Shuffle \cite{7780576}. In application, reshaped versions of $\bar{\mathbf{W}}^l$ (denoted $\bar{\mathbf{W}}^{l,g}$ for sub-pixel group $g$) are applied to neighborhoods $\Omega_{\bar{K}}$ in the original high-level feature map $\mathbf{Y}^{l+1}$:
\begin{equation}
\label{eq:acsm_apply}
\tilde{\mathbf{Y}}_{i,j}^{l+1,g} = \sum_{p,q \in \Omega_{\bar{K}}} \bar{W}_{i,j}^{l,g,p,q} \cdot \mathbf{Y}_{i+p, j+q}^{l+1}.
\end{equation}
Finally, the resulting feature groups $\{\tilde{\mathbf{Y}}^{l+1,g}\}_{g=1}^4$ are then rearranged via Pixel Shuffle to obtain the final upsampled and adaptively smoothed content feature $\tilde{\mathbf{Y}}^{l+1}$:
\begin{equation}
\label{eq:acsm_shuffle}
\tilde{\mathbf{Y}}^{l+1} = \text{PixelShuffle}(\{\tilde{\mathbf{Y}}^{l+1,g}\}_{g=1}^4).
\end{equation}
This process effectively smooths the content representation, enhancing intra-shadow consistency crucial for high-quality shadow removal.

\subsubsection{Texture-Boundary Recuperation Module (TBRM)}

To address the issue where downsampling operations inevitably discard high-frequency information, leading to the loss of fine textures and sharp boundaries in deeper feature maps, we designed the TBRM. This module aims to recover these crucial details by enhancing the high-frequency components inherent in the shallow, high-resolution feature map ($\mathbf{X}^l$). To this end, the core mechanism of TBRM is to predict spatially-variant high-pass filter kernels $\hat{\mathbf{W}}^l$ (of size $\hat{K} \times \hat{K}$), with the prediction based on local contextual features ($Z''$) and utilizing a filter inversion technique \cite{846888}: initial weights ( 
$\hat{\mathbf{V}}^l$) predicted by a convolutional layer are first transformed into low-pass weights via Softmax, then subtracted from an identity kernel $\mathbf{E}$ to yield the high-pass filters:
\begin{equation}
\label{eq:tbrm_invert}
\hat{\mathbf{W}}_{i,j}^l = \mathbf{E} - \text{Softmax}(\hat{\mathbf{V}}_{i,j,:}^l).
\end{equation}
Subsequently, these adaptive high-pass filters $\hat{\mathbf{W}}^l$ are applied to the high-resolution feature map $\mathbf{X}^l$ (or its processed version $\mathbf{X}'^l$) to extract high-frequency information related to textures and edges:
\begin{equation}
\label{eq:tbrm_apply}
\text{HF}(\mathbf{X}'^l)_{i,j} = \sum_{p,q \in \Omega_{\hat{K}}} \hat{W}_{i,j}^{l,p,q} \cdot \mathbf{X}'^l_{i+p, j+q}.
\end{equation}
Finally, this extracted high-frequency component $\text{HF}(\mathbf{X}'^l)$, which represents the details needed for recuperation, is added back to $\mathbf{X}'^l$ via a residual connection. This step yields a detail-enhanced feature map $\tilde{\mathbf{X}}^l$ with recuperated textures and sharpened boundaries:
\begin{equation}
\label{eq:tbrm_residual}
\tilde{\mathbf{X}}^l = \mathbf{X}'^l + \text{HF}(\mathbf{X}'^l).
\end{equation}
In the context of shadow removal, TBRM is vital for restoring the fine textural details often obscured by shadows and for sharpening the transitions at shadow boundaries, contributing significantly to the fidelity and visual quality of the final shadow-free image.

\subsubsection{Component Integration within DFB}
\label{subsubsec:dfb_integration}
As illustrated in Figure \ref{fig:densefuse}, DFB structurally integrates the outputs from its complementary modules. Before the final fusion, to ensure the quality of cross-scale information integration, both the detail-enhanced high-resolution path (from TBRM) and the consistency-focused low-resolution path (from ACSM) are processed through CARAFE \cite{wang2019carafe} modules. As an advanced content-aware feature reassembly technique, CARAFE optimizes the feature maps by dynamically generating upsampling or recombination kernels based on content, thereby ensuring enhanced spatial precision and detail preservation during this critical stage. After adaptive processing by ACSM/TBRM and high-quality reassembly by CARAFE, these two feature streams (smooth base vs. structural details) are then effectively fused, typically via element-wise addition. This structured approach, where different feature components (smooth base vs. structural details) are adaptively processed based on shadow degradation characteristics before recombination, allows DFB to generate a superior fused representation compared to standard methods. This optimized feature map is then passed to the subsequent Transformer Block in the decoder for final contextual refinement.

\paragraph{Training Objective} The model is trained using the Charbonnier loss \cite{zamir2020learning} to supervise the consistency between the estimated shadow-free image $\hat{\mathbf{I}}_f$ and the ground-truth shadow-free image $\mathbf{I}_f$:
\begin{equation}
\label{eq:loss}
\mathcal{L}_{\text{Charbonnier}} = \sqrt{||\mathbf{I}_f - \hat{\mathbf{I}}_f||^2 + \epsilon^2},
\end{equation}
where $\epsilon$ is a small constant (e.g., $10^{-3}$) for numerical stability.

\section{Experiments Results}

\begin{table*}[t]
\small
\center
\begin{tabular}{l c c c c c c c c c c}
\hline
\multirow{2}{*}{Method} & \multirow{2}{*}{Venue} & \multirow{2}{*}{Run-time (ms)} & \multicolumn{2}{c}{ISTD Dataset} & \multicolumn{2}{c}{ISTD+ Dataset} & \multicolumn{2}{c}{SRD Dataset} & \multicolumn{2}{c}{WSRD+ Dataset} \\ \cmidrule(lr){4-5}\cmidrule(lr){6-7}\cmidrule(lr){8-9}\cmidrule(lr){10-11}
& & & {PSNR$\uparrow$} & {SSIM$\uparrow$} & {PSNR$\uparrow$} & {SSIM$\uparrow$} & {PSNR$\uparrow$} & {SSIM$\uparrow$} & {PSNR$\uparrow$} & {SSIM$\uparrow$} \\  
\hline
DSC~\cite{hu2019direction} & TPAMI 2019 & --- & 29.00 & 0.944 & 25.66 & 0.956 & 29.05 & 0.940 & --- & --- \\ 
DHAN~\cite{cun2020towards} & AAAI 2020 & --- & 29.11 & 0.954 & 25.66 & 0.956 & 30.74 & 0.958 & 22.39 & 0.796\\ 
Fu et al.~\cite{fu2021auto} & CVPR 2021 & --- & 26.30 & 0.835 & 28.40 & 0.846 & 28.52 & 0.932 & 21.66 & 0.752\\
BMNet~\cite{zhu2022bijective} & CVPR 2022 & --- & 28.53 & 0.952 & 32.22 &  \cellcolor{colorTrd}{0.965} & 28.34 & 0.943 & 24.75 & 0.816\\ 
TBRNet~\cite{liu2023shadow} & TNNLS 2023 & --- & 28.77 & 0.928 & 31.91 & 0.964 & 31.83 & 0.953 & --- & --- \\ 
ShadowFormer~\cite{shadowformer} & AAAI 2023 & 43.7 & 29.90 & \cellcolor{colorTrd}{0.960} & 31.39 & 0.946 & 30.58 & 0.958 & 25.44 & 0.820\\ 
DMTN~\cite{liu2023decoupled} & TMM 2023 & 82.6 & 29.05 & {0.956} & 31.72 & 0.963 & 32.45 & 0.964 & --- & --- \\
ShadowDiffusion~\cite{guo2023shadowdiffusion} & CVPR 2023 & 506.9 & 30.09 & 0.918 & 31.08 & 0.950 & 31.91 & 0.968 & --- & --- \\ 
ShadowRefiner~\cite{dong2024shadowrefiner} & CVPRW 2024 & --- & --- & --- & --- & --- & --- & --- & 26.04 & \cellcolor{colorTrd}{0.827}\\ 
OmniSR \cite{omnisr} & AAAI 2025 & 120.1 & \cellcolor{colorSnd}{30.45} & \cellcolor{colorSnd}{0.964} & \cellcolor{colorTrd}{33.34} & \cellcolor{colorSnd}{0.970} & \cellcolor{colorTrd}{32.87} & \cellcolor{colorFst}{0.969} & \cellcolor{colorTrd}{26.07} & \cellcolor{colorSnd}{0.835}  \\
StableShadowDiffusion \cite{xu2024detailpreservinglatentdiffusionstable} & CVPR 2025 & 452.8 & --- & --- & \cellcolor{colorFst}{35.19} & \cellcolor{colorSnd}{0.970} & \cellcolor{colorFst}{33.63} & \cellcolor{colorTrd}{0.968} & \cellcolor{colorSnd}{26.26} & \cellcolor{colorTrd}{0.827}  \\
\hline
DenseSR (Ours) & --- & 124.6 & \cellcolor{colorFst}{30.64} & \cellcolor{colorFst}{0.976} & \cellcolor{colorSnd}{33.98} & \cellcolor{colorFst}{0.974} & \cellcolor{colorSnd}{33.45} & \cellcolor{colorFst}{0.970} & \cellcolor{colorFst}{26.28} & \cellcolor{colorFst}{0.838}  \\
\hline
\hline
Fu et al.~\cite{fu2021auto} + GM & CVPR 2021 & ---  & 27.19 & 0.945 & 29.45 & 0.861 & 29.24 & 0.938 & --- & --- \\ 
Zhu et al.~\cite{zhu2022efficient} + GM & AAAI 2022 & --- & 29.85 & 0.960 & --- & --- & 32.05 & 0.965 & --- & --- \\
BMNet~\cite{zhu2022bijective} + GM & CVPR 2022 & --- & 30.28 & 0.959 & 33.98 & \cellcolor{colorTrd}{0.972} & 31.97 & 0.965 & --- & --- \\ 
ShadowFormer~\cite{shadowformer} + GM & AAAI 2023 & 45.1 & \cellcolor{colorTrd}{32.21} & \cellcolor{colorTrd}{0.968} & \cellcolor{colorSnd}{35.46} & 0.971 & {32.90} & 0.958 & --- & --- \\ 
DMTN~\cite{liu2023decoupled} + GM & TMM 2023 & 84.1 & 30.42 & 0.965 & 33.68 & 0.971 & 33.77 & 0.968 & --- & --- \\
ShadowDiffusion~\cite{guo2023shadowdiffusion} + GM & CVPR 2023 & 523.1 & \cellcolor{colorFst}{32.33} & \cellcolor{colorSnd}{0.969} & \cellcolor{colorFst}{35.72} & 0.969 & \cellcolor{colorFst}{34.73} & \cellcolor{colorTrd}{0.970} & --- & --- \\
OmniSR \cite{omnisr} + GM & AAAI 2025 & 122.3 & 31.56 & 0.965 & 34.20 & \cellcolor{colorSnd}{0.973} & \cellcolor{colorTrd}{34.56} & \cellcolor{colorSnd}{0.977} & --- & --- \\
\hline
DenseSR (Ours) + GM & --- & 126.2 & \cellcolor{colorSnd}{32.14} & \cellcolor{colorFst}{0.970} & \cellcolor{colorTrd}{34.64} & \cellcolor{colorFst}{0.974} & \cellcolor{colorSnd}{34.67} & \cellcolor{colorFst}{0.978} & --- & --- \\
\hline
\end{tabular}
\caption{\textbf{Quantitative comparisons on ISTD, ISTD+, SRD, and WSRD+ datasets.} Best results are highlighted as \colorbox{colorFst}{1st}, \colorbox{colorSnd}{2nd} and \colorbox{colorTrd}{3rd}. +GM: using ground-truth shadow masks.}
\vspace{-0.2cm}
\label{tab:comparison_ISTD}
\end{table*}

 
\begin{table}[t]
\setlength\tabcolsep{4pt}
\small
\center
\begin{tabular}{l c c c}
\hline
\multicolumn{3}{c}{INS Dataset}\\
\hline
\multirow{2}{*}{Method} & INS testing & Real testing \\
\cmidrule(lr){2-2}\cmidrule(lr){3-3} & {PSNR$\uparrow$}/{SSIM$\uparrow$} & {PSNR$\uparrow$}/{SSIM$\uparrow$} \\  
\hline
DHAN~\shortcite{cun2020towards} & 27.84/0.963 & 35.05/0.993 & \\ 
Fu et al.~\shortcite{fu2021auto} & 27.91/0.957 & 36.64/0.994 & \\ 
BMNet~\shortcite{zhu2022bijective} & 27.90/0.958 & 36.65/0.994 & \\ 
ShadowFormer~\shortcite{shadowformer} & 28.62/0.963 & \colorbox{colorTrd}{36.99/0.994} & \\ 
DMTN~\shortcite{liu2023decoupled} & {28.83/0.969} & 35.83/0.993 & \\
ShadowDiffusion~\shortcite{guo2023shadowdiffusion} & \colorbox{colorTrd}{29.12/0.966} & {36.91/0.994}
 &\\
OmniSR~\shortcite{omnisr} & \colorbox{colorSnd}{30.38/0.973} & \colorbox{colorSnd}{38.34/0.995}
 &\\
Ours & \colorbox{colorFst}{30.64/0.981} & \colorbox{colorFst}{38.62/0.996} & \\ 
\hline
\end{tabular}
\caption{\textbf{Quantitative comparisons on the INS dataset and real captured images \cite{omnisr}.} Best results are highlighted as \colorbox{colorFst}{1st}, \colorbox{colorSnd}{2nd} and \colorbox{colorTrd}{3rd}.}
\label{tab:comparison_INS}\vspace{-0.6cm}
\end{table}

\subsection{Implementation Details}

We conducted our experiments on ISTD~\cite{wang2018stacked}, ISTD+~\cite{le2019shadow}, SRD~\cite{qu2017deshadownet}, WSRD+~\cite{vasluianu2023wsrd}), and the INS \cite{omnisr} dataset. We evaluated images with a resolution of $256 \times 256$ by random cropping, following previous methods \cite{fu2021auto,le2020shadow,shadowformer,omnisr}. We report results using the commonly used metrics, including Peak Signal-to-Noise Ratio (PSNR) and the Structure Similarity Index Measure (SSIM). For the WSRD+ \cite{vasluianu2023wsrd} dataset, since it does not provide testing data, we used its evaluation data and the evaluation code provided by the NTIRE 2024 Image Shadow Removal Challenge~\cite{vasluianu2024ntire} for comparison. Our model is trained on a GPU server with four GeForce RTX 4090 GPUs using PyTorch 2.0.1 with CUDA 11.7. The batch size and training epoch are set to 3 and 1400, with DDP and AMP training for computational efficiency. We employ the AdamW optimizer \cite{Kingma2015AdamAM} with standard beta parameters ($\beta_1=0.9,\beta_2=0.999$) and an epsilon value of $1 \times 10^{-8}$ for optimization. The initial learning rate is set to $2\times10^{-4}$ and adjusted using a cosine annealing scheduler, configured with a cycle length of 10 epochs and a minimum learning rate of $5 \times 10^{-5}$. Standard data-augmentation strategies, such as random flipping and rotation, are used during training stage.

\begin{figure*}[t!]
\centering
\includegraphics[width=0.95\linewidth]{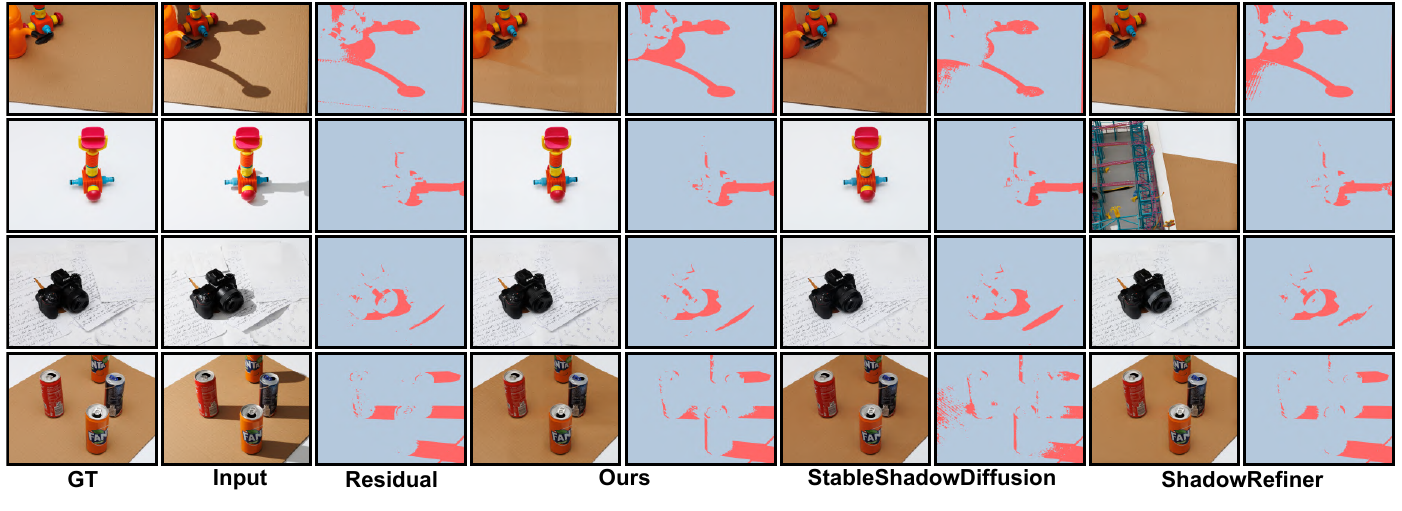}
\vspace{-0.2cm}
\caption{Results visualization with SOTA shadow removal methods, illustrating reconstruction quality of our method on WSRD+ dataset \cite{vasluianu2024ntire}. Comparisons with StableShadowDiffusion \cite{xu2024detailpreservinglatentdiffusionstable} and shadowRefiner \cite{dong2024shadowrefiner}. The residual images are computed by the consistent binary thresholding setting, demonstrating the our method's superiority in details and boundary refinement and addressing ambiguity.}
\vspace{-0.4cm}
\label{fig:comparison} 
\end{figure*}

\begin{figure}[t!]
\centering
\includegraphics[width=1.0\linewidth]{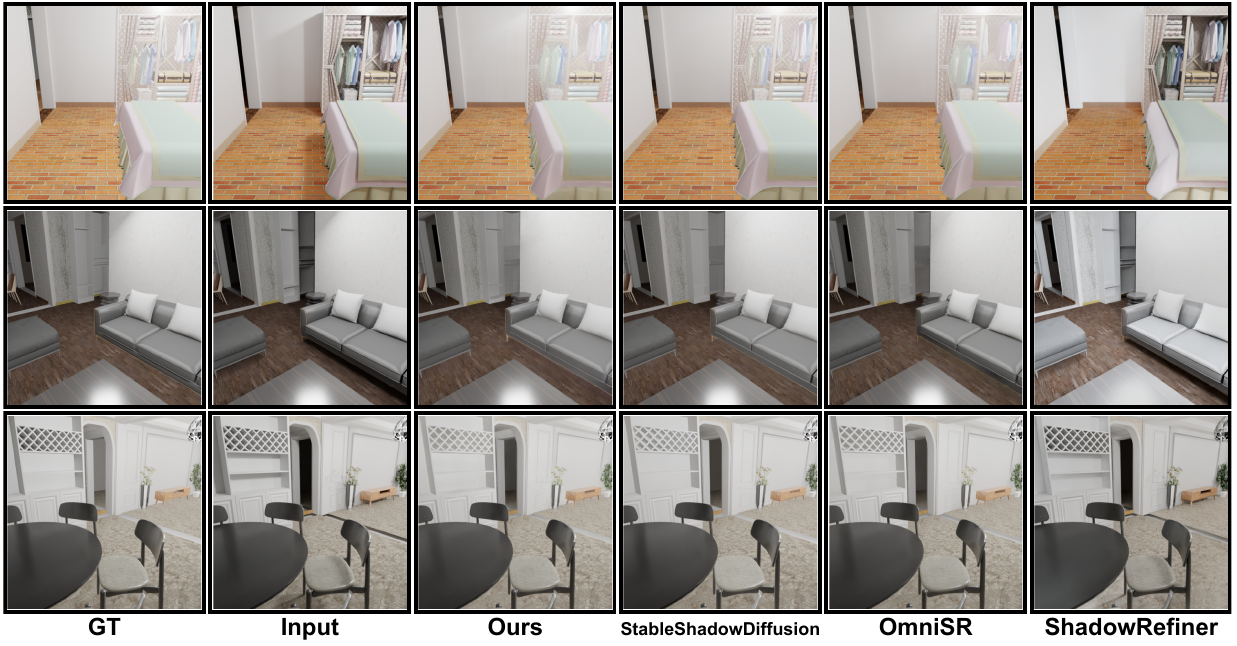}
\caption{Real-world testing data comparisons. For the real captured testing data, our method excels in removing complex indirect shadows and boundary sharpness. (zoom in for better views)
}
\label{fig:ins_real_testing} 
\end{figure}

\subsection{Performance Comparisons}
We compare our method with several state-of-the-art single-image shadow removal methods, including the DSC~\cite{hu2019direction}, DHAN~\cite{cun2020towards}, Fu et al.~\cite{fu2021auto}, Zhu et al.~\cite{zhu2022efficient}, BMNet~\cite{zhu2022bijective}, ShadowFormer~\cite{shadowformer}, DMTN~\cite{liu2023decoupled}, ShadowDiffusion~\cite{guo2023shadowdiffusion} ShadowRefiner~\cite{dong2024shadowrefiner}, OmniSR~\cite{omnisr}, and StableShadowDiffusion~\cite{xu2024detailpreservinglatentdiffusionstable}, as shown in Tables \ref{tab:comparison_ISTD} and \ref{tab:comparison_INS}. The qualitative results are presented in Figure \ref{fig:comparison} and \ref{fig:ins_real_testing}. All comparisons use the results reported in the original papers or the original authors' implementations and hyperparameters. Furthermore, we present a comparison of inference time with the size of a $640 \times 480$ image. Due to the involvement of the pretrained network (e.g. Depth-Anything-V2~\cite{depthganythingv2} and the DINO-V2 network~\cite{dinov2}), our method has higher computational complexity compared to lightweight methods like ShadowFormer~\cite{shadowformer}. However, our method is faster than diffusion-based methods such as ShadowDiffusion~\cite{guo2023shadowdiffusion} and StableShadowDiffusion~\cite{xu2024detailpreservinglatentdiffusionstable}.


As shown in Table~\ref{tab:comparison_ISTD}, our method achieves competitive PSNR and SSIM scores on ISTD, ISTD+, SRD, and INS datasets without GT shadow masks. Even when compared with other methods using GT shadow masks (these methods present their results using GT shadow masks provided by the dataset as the standard input for evaluation, which are not available in real-world applications), our approach, which does not rely on such masks, obtains the second-best results on the ISTD dataset, surpassed only by ``ShadowDiffusion~\cite{guo2023shadowdiffusion} + GM'' and ``ShadowFormer \cite{shadowformer} + GM''. When used with GT masks as network's input, our method also demonstrates a significant improvement in PSNR performance on the ISTD, ISTD+, and SRD datasets. Notably, despite its desired performance, StableShadowDiffusion \cite{xu2024detailpreservinglatentdiffusionstable} needs multiple-stage refinement, significantly costing more computational resources and complexity.


As demonstrated in Figure \ref{fig:comparison}, our method also outperforms other methods evaluated on  the WSRD+ \cite{vasluianu2024ntire} dataset, including ShadowRefiner~\cite{dong2024shadowrefiner} and StableShadowRefiner~\cite{xu2024detailpreservinglatentdiffusionstable}. The relatively low PSNR scores for all methods on the WSRD+ dataset \cite{omnisr} can be attributed to exposure differences between the input and ground-truth images. Concretely, the peer methods still struggle to eliminate shadows in these areas completely in indoor scenes. This limitation may be attributed to these methods lacking explicit adaptive content-aware smoothing and detail-preserving during feature propagation and fusion. We provide additional results in the supplementary material.

\begin{figure}[t!]
\centering
\includegraphics[width=1.0\linewidth]{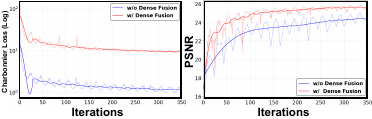}
\vspace{-0.4cm}
\caption{Optimization curve between with/without the proposed DFB, evaluated on WSRD+ \cite{vasluianu2023wsrd} dataset.}
\vspace{-0.45cm}
\label{fig:loss} 
\end{figure}

Evidence of the DFB's benefit is presented in Figure \ref{fig:loss}'s optimization curves, which illustrate that the DFB's specific mechanism—using ACSM for content consistency and TBRM for detail recovery before adaptive fusion—enables the model  to learn more effectively, achieving significantly lower training loss and higher validation PSNR compared to the model lacking this adaptive fusion capability. This highlights how the DFB's design, which balances content smoothing with detail refinement, directly translates to better optimization and improved shadow removal performance.

\begin{table}[t]
\setlength\tabcolsep{4pt}
\small
\center
\begin{tabular}{c c c c}
\hline
&\multicolumn{3}{c}{Dataset}\\\hline
\multirow{2}{*}{} & INS Testing & Real captured & WSRD+ \\
\cmidrule(lr){2-2}\cmidrule(lr){3-3}\cmidrule(lr){4-4}Configuration& {PSNR$\uparrow$}/{SSIM$\uparrow$} & {PSNR$\uparrow$}/{SSIM$\uparrow$} & {PSNR$\uparrow$}/{SSIM$\uparrow$} \\
\hline
Full & \textbf{30.64/0.981} & \textbf{38.64/0.996} & \textbf{26.28/0.838} \\
W/o depth & 29.95/0.973 & 38.03/0.995 & 25.73/0.828 \\ 
W/o normal & 29.32/0.967 & 37.75/0.995 & 25.80/0.829 \\
W/o DINO & 29.31/0.966 & 37.06/0.994 & 23.31/0.797 \\
W/o DFB & 30.38/0.973 & 38.34/0.995 & 26.07/0.835 \\
W/o ACSM & 30.52/0.974 & 38.48/0.995 & 26.11/0.836 \\
W/o TBRM & 30.49/0.972 & 38.43/0.994 & 26.12/0.836 \\
\hline
\end{tabular}
\caption{\textbf{Ablation studies for all modules.} W/o depth: only RGB input. W/o DINO: without DINO feature. W/o DFB: without adaptive feature fusion.}
\vspace{-1cm}
\label{tab:ablation}
\end{table}

\subsection{Ablation Study}
To validate our model designs, we conducted ablation studies on the proposed semantic and geometric attention weights, depth concatenation, and DINO feature concatenation. The ``INS testing'' and ``real captured'' are trained on the INS training dataset. The ``WSRD+'' is trained and evaluated using the WSRD+ dataset \cite{vasluianu2024ntire}. To validate the effectiveness of key designs in the DenseSR model, we conducted ablation studies summarized in Table \ref{tab:ablation}. The analysis reveals that removing depth (W/o depth) or normal maps (W/o normal) degrades performance, confirming the value of geometric cues, while removing DINO semantic features (W/o DINO) causes a significant performance drop, highlighting the critical role of high-level semantic priors in shadow identification and ambiguity reduction. Concurrently, replacing the DFB with standard fusion (W/o DFB) also leads to lower performance, demonstrating the superiority of our adaptive dense fusion strategy. Furthermore, ablating ACSM (W/o ACSM) or TBRM (W/o TBRM) individually within the DFB results in slight performance decreases, validating the respective contributions of the content smoothing module to consistency and the texture-boundary recuperation module to detail fidelity. In summary, the ablation results conclusively demonstrate that DenseSR's superior performance stems from the synergistic interplay between multi-modal prior integration (especially DINO and depth) and the adaptive component processing (ACSM and TBRM) within the DFB.

\vspace{-0.2cm}
\section{Conclusion}
To address inconsistent content restoration and boundary blurring in single-image shadow removal—caused by ambiguity, complex lighting, and non-uniform feature degradation, in this paper, we introduced the DenseSR framework. From the dense prediction perspective, DenseSR leverages prior-guided attention for spatial-scene understanding and ambiguity reduction, and employs the innovative DFB with adaptive feature fusion with content-filtering manner to overcome standard fusion limitations for high-fidelity content restoration. The ACSM ensures smooth content consistency within restored shadows, while the TBRM crucially recuperates fine textures and sharpens boundaries. Experiments validate DenseSR's SOTA performance on multiple benchmarks and its effectiveness in handling complex shadows.
\\

\textbf{Acknowledgements.} We thank to National Center for High performance Computing (NCHC) of National Applied Research Laboratories (NARLabs) in Taiwan for providing computational and storage resources.

\bibliographystyle{ACM-Reference-Format}
\bibliography{reference}
\clearpage

\renewcommand\thetable{\Alph{table}}
\renewcommand\thefigure{\Alph{figure}}

\vspace{1em}
\noindent\textbf{\LARGE Supplementary Material}\\[0.5em]
\noindent\textbf{DenseSR: Image Shadow Removal as Dense Prediction}
\vspace{1em}


\noindent{In the supplementary material, we present the following:}
\begin{itemize}
    \item Network Hyperparameter Settings
    \item Data Loading and Preprocessing
    \item More Visual Comparisons
\end{itemize}

\section{Network Hyperparameter Settings}  
\label{sec:imp}

\subsection{DenseSR}
A key architectural modification involves the strategic integration of three specialized Dense Fusion Blocks ($\text{DFB}_1$, $\text{DFB}_2$, and $\text{DFB}_3$) at the decoder's fusion points, replacing conventional feature fusion operations to facilitate enhanced cross-level feature integration, as detailed in Section \ref{dfb}. The network parameters are initialized with a normal distribution $\mathcal{N}(0,0.2)$. We configure DenseSR with a base embedding dimension of 32. The architecture comprises seven primary processing stages (three encoder stages, one bottleneck, and three decoder stages), each consistently utilizing a depth of 2, signifying that they are composed of two stacked transformer blocks, CAB and SIM.
The number of attention heads within these respective blocks varies across the stages, following the sequence 1, 2, 4, 16, 8, 4, and 2, corresponding to the three encoder stages, the bottleneck, and the three decoder stages. The window size of 16 is employed for all window attention with window shifting stride $16//2$. Furthermore, the MLP expansion ratio within each Transformer block is set to 4.


\subsection{Dense Fusion Block}
\label{dfb}
The DFB module composed primarily of the Texture-Boundary Recuperation Module (TBRM) and the Adaptive Content Smoothing Module (ACSM), accepts two primary inputs: a high-resolution feature map (hr\_feat) and a low-resolution feature map (lr\_feat). Both input feature maps initially undergo channel compression via dedicated $1 \times 1$ convolutions. These convolutional layers reduce the channel dimensions of both hr\_feat and lr\_feat to an intermediate dimension of 64, as specified by the compressed channels hyperparameter, to subsequent adaptive kernel prediction against computational efficiency.

Following this compression, the DFB predicts spatially-variant kernels based on the derived context. Specifically, we set the low-pass kernel to $5 \times 5$ in ACSM. The corresponding kernel generator predicts raw weights, outputting a feature map where the channel dimension is proportional to the squared kernel size. Within the crucial kernel normalizer step, these raw weights are first reshaped to isolate the kernel dimension $K\times K$. A channel-wise softmax is then applied across this $K\times K$ dimension, transforming the raw predictions into normalized weights that sum to one. After optional Hamming windowing and re-normalization, the output is reshaped again to yield the final, stable $5 \times 5$ spatially-variant low-pass filter kernels (mask\_lr) ready for application via CARAFE \cite{wang2019carafe} module. This larger kernel size facilitates effective feature smoothing and captures the broader spatial context necessary for ensuring content consistency.

Concurrently, for TBRM, a smaller kernel of $3 \times 3$ is utilized. Its generator similarly predicts initial raw weights, with channels proportional to the squared kernel size. These weights undergo the identical kernel normalizer process involving reshaping and channel-wise softmax across the ${K'}\times{K'}$ dimension, optionally modulated by a Hamming window \cite{hamming}. This step critically produces normalized, stable intermediate kernels (mask\_hr), which are conceptually akin to low-pass filters before the final transformation. The effective $3 \times 3$ high-pass filter required by TBRM is then derived implicitly through filter inversion (conceptually, subtracting mask\_hr from an identity kernel). This compact $3 \times 3$ kernel is better suited for precisely identifying and enhancing highly localized textural details and sharp edges inherent in the high-frequency components targeted by TBRM.

\subsection{Data Loading and Preprocessing}
The proposed DenseSR needs the four inputs: (1) RGB image, (2) Depth map, (3) Normal map, and (4) Semantic feature map. First, Depth-Anything-V2 \cite{depthganythingv2} and DINO-V2 \cite{dinov2} are utilized to extract external features. Note that using pre-trained models like these is commonly used in the field recently. Afterwards, the normal map can be obtained by normal estimation using the depth map with camera intrinsics. Concretely, this conversion utilizes the camera's field of view (FOV, specified as 60 degrees in the implementation) and the image dimensions $H$ and $W$ to first calculate the camera's focal length $f$ and principal point $c_x, c_y$ using:
\begin{equation}
    f   = \frac{W}{2 \tan(\text{FOV}_{\text{radians}} / 2)};\quad
    c_x = \frac{W - 1}{2};\quad
    c_y = \frac{H - 1}{2}
\end{equation}
where $\text{FOV}_{\text{radians}} = \text{FOV}_{\text{degrees}} \times \frac{\pi}{180}$. Then, for each pixel coordinate $x, y$ with its corresponding depth value $z=\text{depth}[y,x]$, the 3D coordinates $(x_{3d},y_{3d},z)$
 are computed using the pinhole camera model equations:
\begin{equation}
    x_{\text{3d}} = \frac{(x - c_x) \times z}{f};\quad
    y_{\text{3d}} = \frac{(y - c_y) \times z}{f}
\end{equation}
These calculated 3D points $(x_{3d},y_{3d},z)$ for all pixels are stacked together to form the final normal map, represented as an array of shape $(H, W, 3)$.

Concurrently, the loaded surface normal map, initially assumed to be in the [0, 1] range often derived from rendering or estimation, undergoes processing to ensure it represents properly normalized 3D vectors suitable for geometric calculations. First, the normal map values $\mathbf{n}_\text{raw}$ are linearly rescaled to the [-1, 1] range:
\begin{equation}
    \mathbf{n}_{\text{rescaled}} = \mathbf{n}_{\text{raw}} \times 2.0 - 1.0    
\end{equation}

Then, each per-pixel normal vector $\mathbf{n}_\text{rescaled}$ is explicitly normalized by calculating the L2 norm of each vector and dividing the vector components by this magnitude:
\begin{equation}
     \mathbf{n}_{\text{normalized}} = \frac{\mathbf{n}_{\text{rescaled}}}{\|\mathbf{n}_{\text{rescaled}}\|_2 + \epsilon}
\end{equation}
where $\epsilon$ is set to $10^{-20}$. The resulting processed normal map contains unit-length vectors representing surface orientations. During training time, we randomly cropped the input image from source data with the size of $256\times256$. In order to fit the resolution of different stages in the base model, we then partition these external data into grid samples. So far, four distinct data modalities are acquired. 

\section{Analysis and Comparisons}
\subsection{Frequency Domain analysis}
\begin{figure}[t!]
\centering
\includegraphics[width=0.65\linewidth]{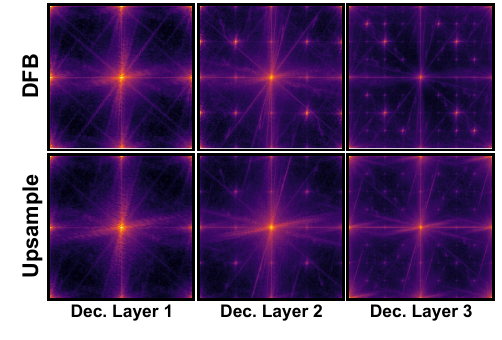}
\vspace{-0.4cm}
\caption{Frequency domain analysis comparing DFB (top row) and standard decoder (bottom row) outputs at corresponding stages. Note the significantly stronger high-frequency components (brighter periphery) in the DFB spectra, indicating enhanced detail and boundary representation.}
\vspace{-0.45cm}
\label{fig:frequency} 
\end{figure}

To compare the standard decoder mechanism and the proposed (DFB), we analyze their respective outputs in the frequency domain. Figure \ref{fig:frequency} visualizes the Fast Fourier Transform (FFT) magnitude spectra derived from feature maps at corresponding decoder layers for both approaches. The top row displays the spectra for DFB's high-resolution outputs, while the bottom row shows those from the baseline 
standard decoder at equivalent stages.

In these spectra, the center (low frequencies) reflects slowly varying components like overall structure, while the periphery (high frequencies) represents rapid changes such as edges, textures, and details, with energy levels indicating component prevalence. This comparison clearly reveals the DFB's strength in details refinement. The DFB outputs (top row) exhibit markedly stronger high-frequency components, directly correlating with its objective of enhancing detail and sharpening boundaries. This enhanced high-frequency energy translates to the sharper edges, clearer textures, and better-preserved details observed spatially in DenseSR's results. Conversely, the standard decoder (bottom row), lacking explicit high-frequency recuperation, shows weaker high-frequency energy, consistent with potentially smoother, less detailed outputs.

\subsection{More Visual Comparisons}
\begin{figure}[t!]
\centering
\includegraphics[width=1.0\linewidth]{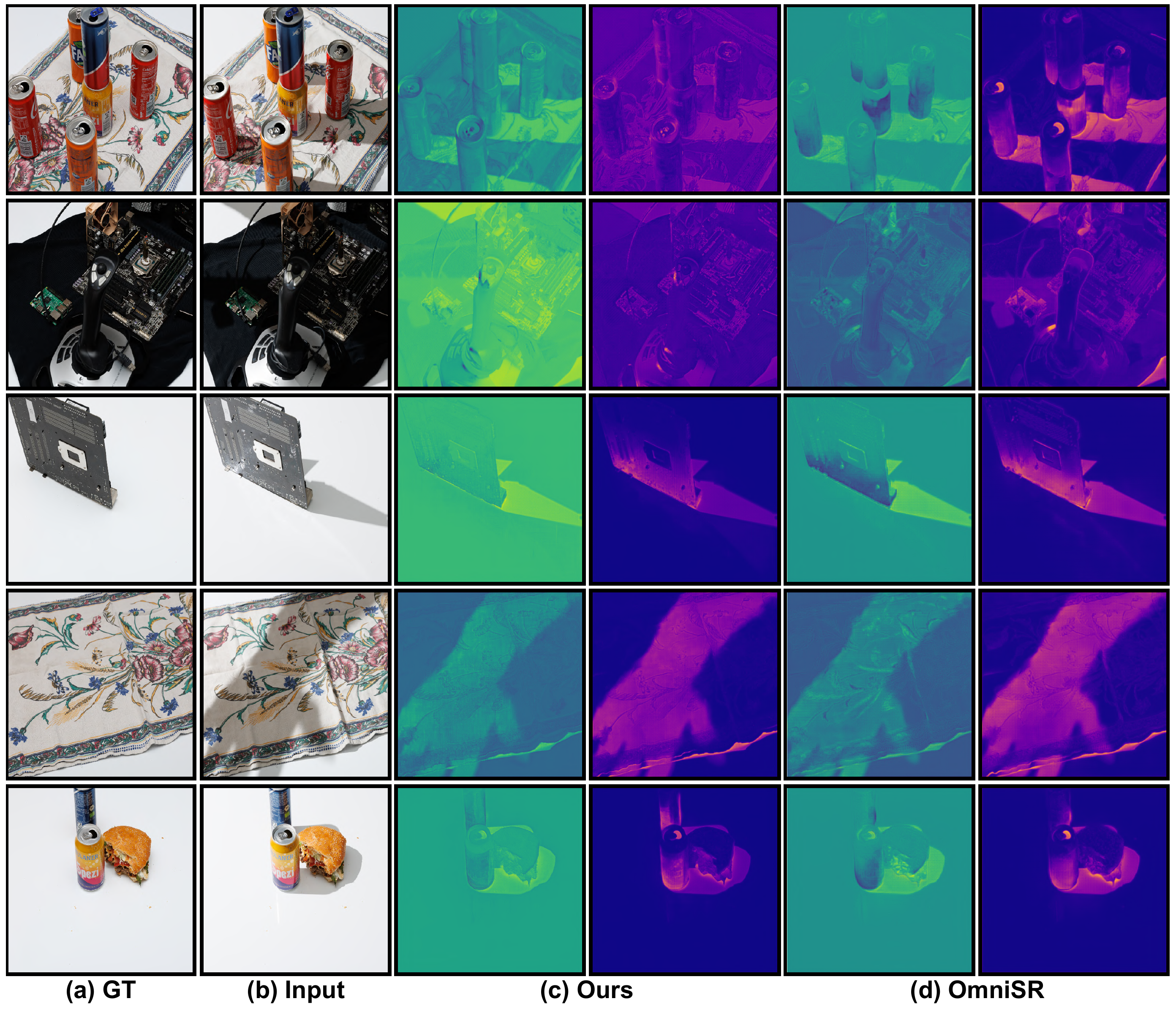}
\vspace{-0.4cm}
\caption{Additional visualizations comparing the mean and variance of feature maps.}
\vspace{-0.45cm}
\label{fig:more_avg_var} 
\end{figure}

\begin{figure}[t!]
\centering
\includegraphics[width=1.0\linewidth]{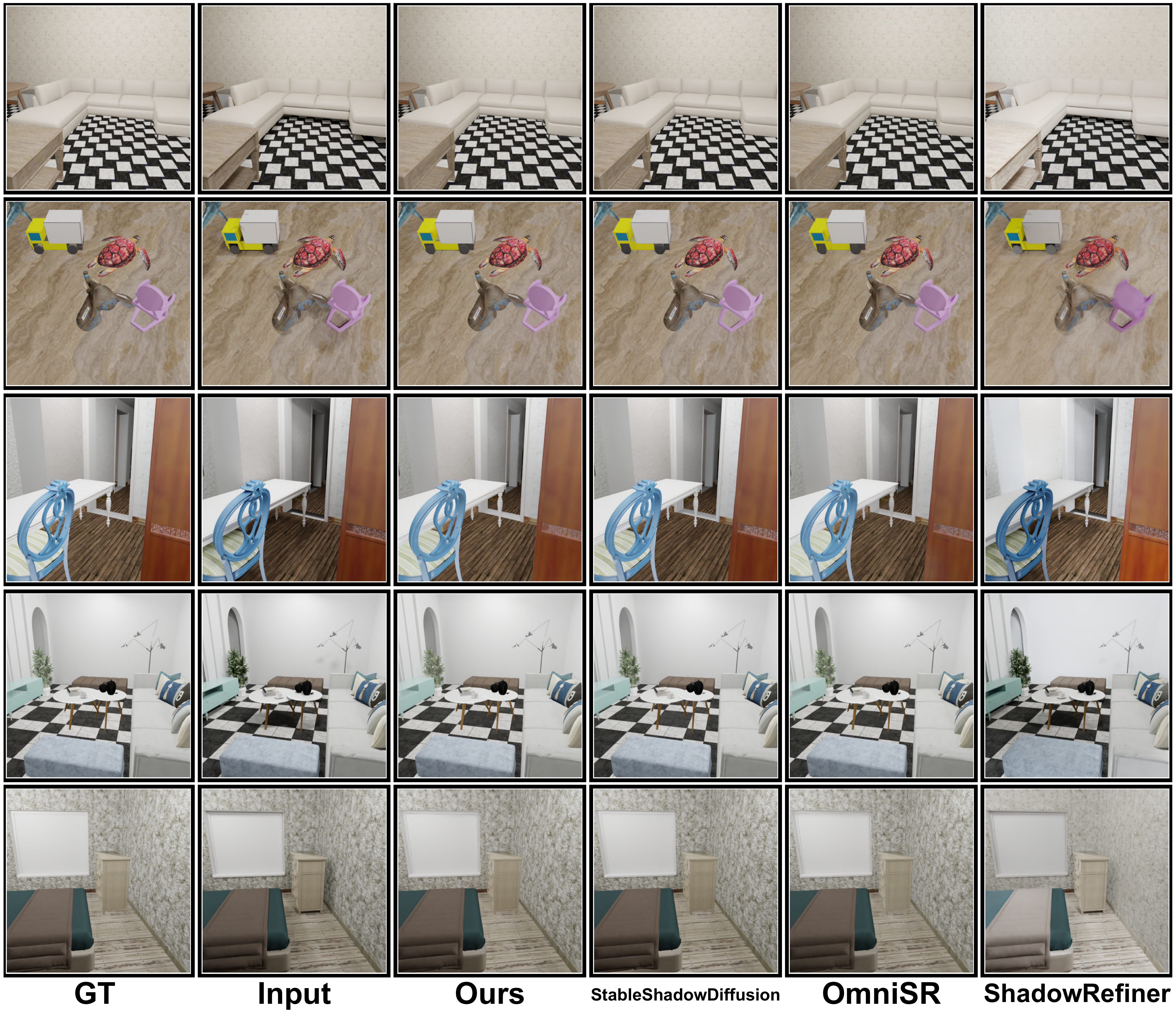}
\vspace{-0.4cm}
\caption{More visual comparison of shadow removal results.}
\vspace{-0.45cm}
\label{fig:more_result_compare} 
\end{figure}

Figure \ref{fig:more_avg_var} presents additional visual comparisons focusing on the mean and variability of features on the WSRD+ \cite{vasluianu2023wsrd} dataset, while Figure \ref{fig:more_result_compare} showcases further examples of our shadow removal results on the INS \cite{omnisr} dataset.

\appendix

\end{document}